# Vec2Vec: A Compact Neural Network Approach for Transforming Text Embeddings with High Fidelity

**Andrew Kean Gao**
Stanford University
gaodrew@stanford.edu

## Abstract:

Since the seminal Word2Vec paper in 2013, vector embeddings have become ubiquitous and powerful tools for many language-related tasks. A leading embedding model is OpenAI's text-ada-002 which can embed up to approximately 6,000 words into a 1,536-dimensional vector. While powerful, text-ada-002 is not open source and is only available via API. Thus, users must have an Internet connection to query their text-ada-002 databases. Additionally, API costs can add up and users are locked in to OpenAI. We trained a simple neural network to convert open-source 768-dimensional MPNet embeddings into text-ada-002 embeddings. We compiled a subset of 50,000 reviews from Stanford's Amazon Fine Foods dataset. We calculated MPNet and text-ada-002 embeddings for each review and trained a simple neural network for 75 epochs. Our model achieved an average cosine similarity of 0.932 on 10,000 unseen reviews in our held-out test dataset. Given the high dimension (1,536) of text-ada-002 embeddings, 0.932 is quite impressive. Finally, we manually assessed the quality of our predicted "synthetic" embeddings for vector search over text-ada-002-embedded reviews. While not as good as real text-ada-002 embeddings, predicted embeddings were able to retrieve highly relevant reviews. Our final model, "Vec2Vec", is lightweight (<80 MB) and fast. Future steps include training a neural network with a more sophisticated architecture and a larger dataset of paired embeddings to achieve greater performance. The ability to convert between and align embedding spaces may be helpful for interoperability, limiting dependence on proprietary models, protecting data privacy, reducing costs, and offline operations.

# Introduction

Embeddings are a powerful technique in natural language processing that allow us to represent texts as vectors in a high-dimensional space [1-2]. These vectors capture the semantic and syntactic relationships between texts, enabling us to perform various tasks such as search, sentiment analysis, language translation, and text classification. Embeddings are typically learned using unsupervised techniques such as word2vec or GloVe, which use large amounts of text data to learn the vector representations of words

[3-4]. The Word2Vec model, developed by Tomas Mikolov and his team at Google in 2013, is a neural network that processes text by taking as its input a large corpus of words and producing a vector space. Each word in the corpus gets assigned a corresponding vector in the space. Words that share common contexts in the corpus are placed in close proximity to one another in the space.

Embeddings have revolutionized the field of natural language processing and have become an essential tool for building state-of-the-art models in various applications [5]. Embeddings have widespread applications in many natural language processing tasks [6-8]. They are often used in search, where one can query the embedding space to find vectors closest to a given vector and return relevant documents. Text similarity and clustering involve comparing the embeddings of different pieces of text and grouping similar ones together.

OpenAI's text-ada-002 is an advanced embedding model that represents large texts as high-dimensional vectors [9]. It is at the top of the leaderboards in various benchmarks and can embed up to approximately 6,000 words into a 1,536-dimensional vector. The model is quite powerful but is proprietary, meaning that it's not open-source and is only available via API. Also, there are rate limits. This has implications on costs, dependency, and requirement of an Internet connection. On the other hand, all-mpnet-base-v2 is an open-source embedding model that is freely available and can be run locally [10]. The creators of all-mpnet-base-v2 fine-tuned Microsoft's MPNet model on 1 billion sentence pairs [11].

In this paper, we train a neural network to learn to convert embeddings generated by an all-mpnet-base-v2 into text-ada-002 embeddings. Neural networks are essentially networks of algorithms aimed at recognizing underlying relationships in a set of data, modeled off of the human brain. They are designed to recognize patterns and are exceptionally effective at processing complex, high-dimensional data such as images, audio, language, or in our case, high-dimensional vectors. Neural networks, with their ability to learn complex mappings and generalize from seen to unseen data, are ideal for our purpose.

In order to train a neural network, a loss function is necessary. In this study, we use cosine similarity. Cosine similarity is a metric used to measure how similar two vectors are, irrespective of their size. It measures the cosine of the angle between two vectors projected in a multi-dimensional space. The smaller the angle, the higher the cosine similarity. In the context of text embeddings, cosine similarity is a good metric as it effectively captures the orientation (direction) of the embeddings, which is more important than their magnitude in high-dimensional spaces. In the context of our research, we calculate the cosine similarity between our predicted embeddings and the real text-ada-002 embeddings, as a sort of accuracy metric.

Let $y_\text{true}$ and $y_\text{pred}$ be two 1536-dimensional embedding vectors, where 'true' refers to the ground truth text-ada-002 vector in the test dataset and 'pred' refers to the vector predicted by Vec2Vec given an all-mpnet-base-v2 vector. The L2 normalization of these vectors is given by:

$$y_{\text{true, norm}} = \frac{y_\text{true}}{\|y_\text{true}\|_2} = \frac{y_\text{true}}{\sqrt{\sum_i (y_\text{true})_i^2}}$$

$$y_{\text{pred, norm}} = \frac{y_\text{pred}}{\|y_\text{pred}\|_2} = \frac{y_\text{pred}}{\sqrt{\sum_i (y_\text{pred})_i^2}}$$

We perform element-wise multiplication on these normalized vectors:

$$z = y_{\text{true, norm}} \odot y_{\text{pred, norm}}$$

Then we take the mean of the resulting vector. Because the vectors are normalized to length 1, we do not need to divide by the product of their lengths because that would be dividing by 1:

$$\mu_z = \frac{1}{N} \sum_i z_i$$

Finally, the cosine similarity loss is the negative of this mean:

$$\text{Cosine similarity loss} = -\mu_z$$

*Figure 1.* Description and formulas of our custom cosine similarity loss function.

The overarching idea of this study is to essentially map one embedding space onto another. One example application of our work is in vector databases. The initial vector database can be created using text-ada-002. At inference/search time, all-mpnet-base-v2 + Vec2Vec can be used to embed the search query instead of text-ada-002.

# Methods:

We retrieved 568,454 Amazon reviews of fine foods from the Stanford Network Analysis Project [12]. The reviews are written by 256,059 unique Amazon users and span 74,258 food products, with an average of 7.65 reviews per product. The median number of words per review is 56. The reviews were posted on Amazon between October 1999 and October 2012. We selected this dataset because it is large, has texts of a manageable size, and because the domain is relatively constrained (food) while still being diverse (dog food, milk, spices, meat, vegetables, instant noodles, canned soups, etcetera).

We downloaded the reviews in CSV format and used Python to preprocess the data. Because the OpenAI text-ada-002 model can only embed up to 8,192 tokens at once, we used the tiktoken package to remove reviews of length greater than 8,000 tokens, if any existed. Next, we randomly sampled 50,000 reviews from the remaining reviews using the Pandas sample method.

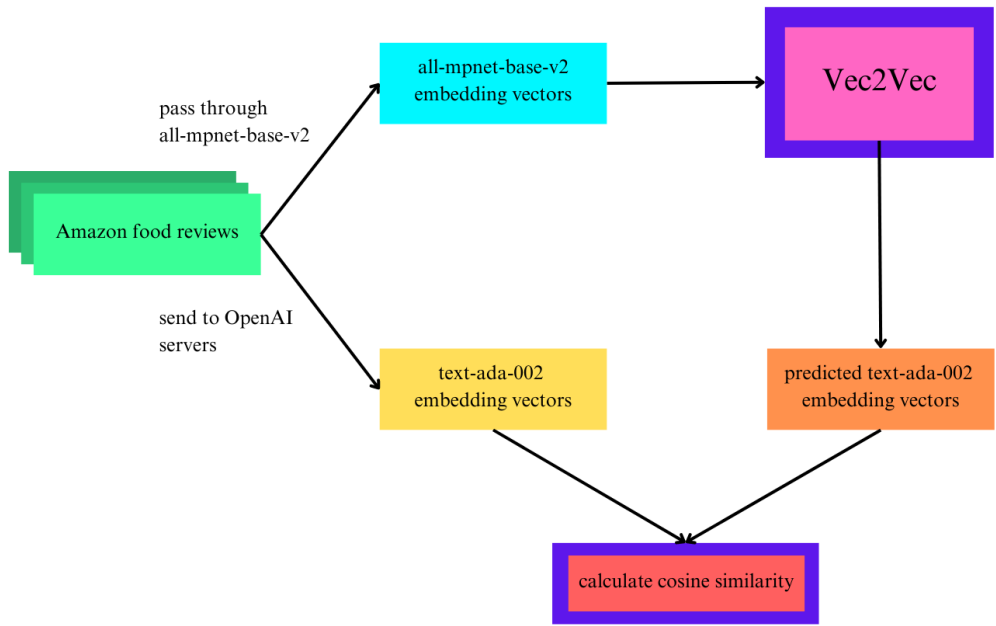

*Figure 2.* Schematic showing the simplified workflow of training the model.

Next, we calculated vector embeddings for these reviews using the open-source model all-mpnet-base-v2 (obtained from Hugging Face) and the OpenAI text-ada-002 model (June 2023 version). We referenced code provided by OpenAI on how to use text-ada-002. In order to obtain 50,000 embeddings from OpenAI in a reasonable timeframe, we used the LightspeedEmbeddings package to implement multithreading and send multiple API requests simultaneously [13]. Also, all-mpnet-base-v2 has a limit of approximately 384 'word pieces' per embedding so we divided any long reviews into chunks of 128 words, embedded the chunks separately, and then computed average embeddings. However, the vast majority of reviews fell under this 128 word threshold so this was not often needed. The all-mpnet-base-v2 model provided a 768-dimensional embedding vector and the text-ada-002 model provided a 1536-dimensional embedding vector, which happens to be exactly twice the size of the former.

We built a simple fully connected sequential neural network using the Tensorflow and Keras libraries in Python [14-15]. The model had three hidden layers with ReLU activation functions and three dropout layers to combat overfitting. We used a custom loss function, which was just a cosine similarity but negative. The input features were the all-mpnet-base-v2 vectors and the output features were the corresponding text-ada-002 vectors. We reserved a random 20% of the data (10,000 reviews) for the final test set. The model did not see any of these reviews during training. We trained it for 75 epochs with a batch size of 32, the Adam optimizer, and a validation split of 20%, using cosine similarity for the loss function. We used cosine similarity in order to encourage the neural network to minimize the difference in angle between its predicted vectors and the ground truth text-ada-002 vectors. Due to the high dimension of the vectors, Euclidean distance would not be appropriate. Additionally, the orientation, rather than magnitude, of embedding vectors is more important.

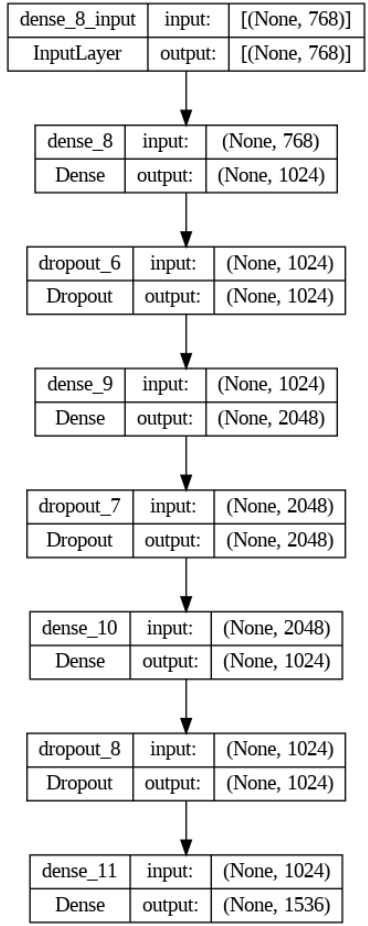

*Figure 3.* Architecture of the Vec2Vec neural network.

To visually assess whether our model was truly accurate, we set up a simple test. We saved the model and chained it with the all-mpnet-base-v2 model. Our test program would solicit a search query from the user. It would first embed the search query using the all-mpnet-base-v2 model. Then, it would send that vector into our custom model and obtain a predicted text-ada-002 embedding. Then, we performed vector search over the real text-ada-002 embeddings in our 10,000-review test set by ranking cosine similarities with our predicted embedding. We repeated the process using a real text-ada-002 embedding for our search query instead of a predicted embedding. While not rigorous, this approach enabled us to quickly verify whether our model was at all useful and able to replace text-ada-002 to some extent.

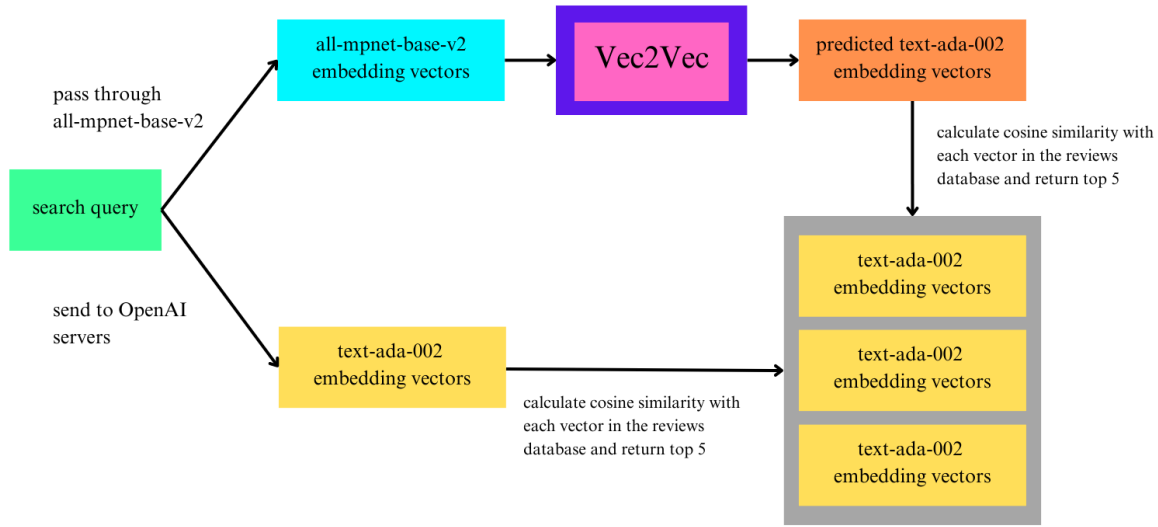

*Figure 4.* Schematic showing how we tested all-mpnet-base-v2 + Vec2Vec against text-ada-002 for retrieving relevant reviews from our 10,000-review dataset.

# Results

After training for 75 epochs, the model achieved a validation loss of -0.00060648. The validation loss decreased smoothly with the training loss. We computed the cosine similarity between each of our 10,000 predicted embeddings and the corresponding true text-ada-002 embeddings. It was 0.932. The maximum possible cosine similarity between any two vectors is 1.

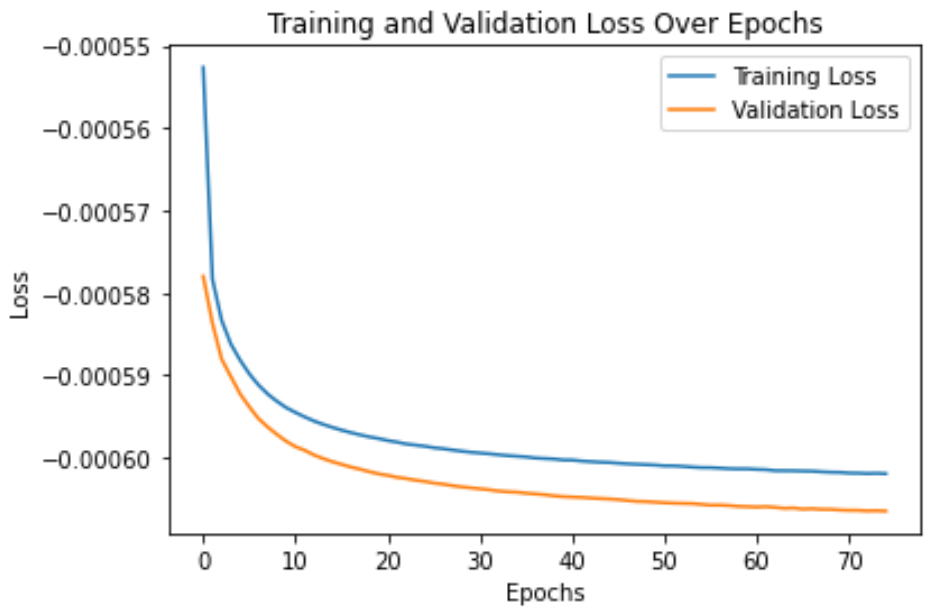

*Figure 5*. Graph showing the decrease in training loss and validation loss over 75 epochs.

The standard deviation in the cosine similarities was 0.0208. The worst cosine similarity was -0.044, an outlier, and the best cosine similarity was 0.977. The vast majority of cosine similarities fell between 0.85 and 0.975.

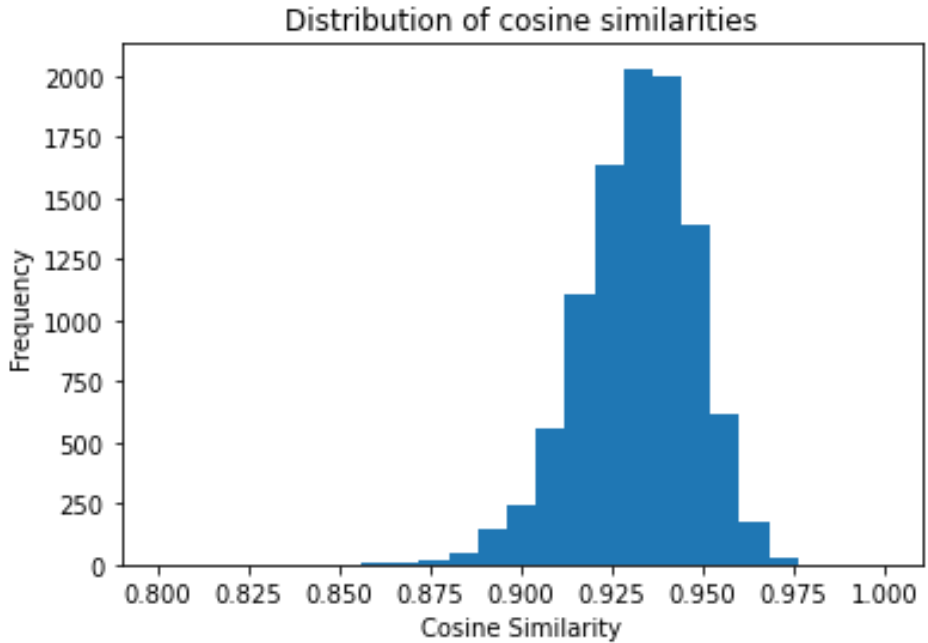

*Figure 6*. Histogram of cosine similarities between each of the 10,000 predicted embedding vectors and 10,000 ground truth text-ada-002 embedding vectors in the held-out test dataset. The average cosine similarity was 0.932.

To manually compare our predicted embeddings versus real text-ada-002 embeddings on performing vector search on a dataset of reviews embedded with text-ada-002, we wrote a program to return the top five best matching reviews to a search query in order. The search query would be embedded with text-ada-002 as well as all-mpnet-base-v2 + Vec2Vec. The search query was first embedded using all-mpnet-base-v2. Then, that 768-dimensional embedding was submitted to the Vec2Vec model to obtain a predicted 1536-dimensional text-ada-002 embedding.

We found that the predicted embeddings performed well, but sometimes worse than real text-ada-002 embeddings. On simple queries, our Vec2Vec embedding returned many of the same reviews as the text-ada-002 embedding. Vec2Vec and all-mpnet-base-v2 worked very well (on par with text-ada-002) on short, semantically simple queries, such as "ramen noodles", "spicy", and "dog food", but performed worse on more complex queries such as "the dog food was expired" and "the wine had notes of lavender". However, text-ada-002 also struggled on these complex queries. For example, it mainly returned reviews about the plant lavender rather than reviews about the taste of wine. We have compiled the different search results in tables below.

| Search | Vec2Vec+ all-mpnet-base-v2 embedding results | text-ada-002 embedding results |
| --- | --- | --- |

| Rank | | |
|---|---|---|
| 1 | Title: Spicy and flavorful; Content: This is a great product. It comes in a nice selection with all levels of extreme heat represented. My hubby likes to get others to try it who think they can handle it. Others always comment the it is hot but it also has great flavor. We like things hot but they also need to taste good!! This variety pack does not disappoint. | Title: Spicy and flavorful; Content: This is a great product. It comes in a nice selection with all levels of extreme heat represented. My hubby likes to get others to try it who think they can handle it. Others always comment the it is hot but it also has great flavor. We like things hot but they also need to taste good!! This variety pack does not disappoint. |
| 2 | Title: So Yummy!!; Content: Not overly hot, but still contains the spice to kick up your tastes buds. Absolutely impressed with this product and will continue to use in the future. Nothing is wouldn't be good on!! | Title: spicy; Content: This was pretty good and goes easily on many things such as hamburgers and hotdogs etc and wouldn't make a terrible salad dressing either. Its not necessarily healthy however. |
| 3 | Title: Nice Heat, Nice Flavor; Content: Very good, one of my new favorite toppings that adds great flavor and just the right amount of heat to any meal. | Title: Spicy; Content: Just licking the dust off of the freshness seal set my mouth on fire! These aren't the sissy hot pepper flakes you find at a pizza shop. |
| 4 | Title: Nice Heat, Nice Flavor; Content: Very good, one of my new favorite toppings that adds great flavor and just the right amount of heat to any meal. | Title: Spicy, but not hot; Content: I liked this one and I was relieved to see there is no MSG included in the ingredients list, unless they're hiding it under an unfamiliar name. It's spicy but not hot spicy, creamy and tastes good. I used it as a dip for chicken strips and think it would make a good dressing for taco salad and in burritos. |
| 5 | Title: The spice will grow!; Content: At first you think that these chips aren't all that Spicy, but the spice will grow on you with each bite. They are really very good! | Title: The spice will grow!; Content: At first you think that these chips aren't all that Spicy, but the spice will grow on you with each bite. They are really very good! |

*Table 1.* The top five search results returned from our text-ada-002 vector database of 10,000 reviews when using all-mpnet-base-v2 + Vec2Vec versus text-ada-002 for the query "spicy". Both models performed well and returned relevant results.

| Search Rank | Vec2Vec+ all-mpnet-base-v2 embedding results | text-ada-002 embedding results |
|---|---|---|
| 1 | Title: Great & yummy noodles, you can't go wrong; Content: My husband and I have eating these since college. 10 years later, we still love it. We usually get them from the asian supermarket, but the price is actually better on Amazon.<br /><br />If you love spicy stuff, you will love this. I don't eat things that are too spicy, so when I make it for myself, I only put in half the spice package. Yummy! | Title: inexpensive meal; Content: The Oriental Flavor Ramen Noodles taste okay but I almost always add ingredients to make the dish more wholesome. I prepare it by adding sesame seed oil, vegetables and water then adding the seasoning and noodles after the water boils. Rough vegetables like chopped onions taste better if they are boiled with the seasoning for 15 minutes prior to adding the noodles. |
| 2 | Title: delicious noodle; Content: I like this noodle so much. Easy to make it, unexpensive, has a lot of flavours. I bet everybody will like this noodle also if they tried. | Title: Pretty awesome, but could be better; Content: Overall, I'd give the noodles themself 5 stars. Super easy and quick to make, and healthier than that cheap junk. This ramen says "I'm not a starving college student anymore, and dammit I still love instant ramen". Or something like that. Anyway... I am withholding one star due to the seasoning packet, that quite frankly, isn't that great, and very bland. I find that |

| | | | |
|---|---|---|---|
| | | | a healthy amount of black pepper, red chili flake, and a few drops of your favorite hot sauce and you're in business. |
| | 3 | Title: It tastes really good.; Content: I have tried many types of noodles and vermicellies and this one by far the best one. It is very easy to prepare (only took me about 20 minutes to get my meal ready). The season is just enough, not too light, not to salty. And did I mention that it tastes awesome? The only drawback is that it's kind of dry. But if you have it with some sauce, it would not be a problem. | Title: A Good Choice; Content: This is a classic ramen choice. Our kids, 3-6 love this. When we're RVing, we have a case of this on hand. A bowl, some water, and 2:30 in the microwave, and we've got our instant meal. We break up the noodles while still sealed in the package. This helps the kids not make a mess when trying to eat them. They don't seem to tire of these; and we're not overly offended by the nature of the just-add-water-product. We each had our ramen-eating-college-days. These also make a great item to have on hand in a desk drawer for the days you forget your lunch, etc. at work. Just make sure you have a pack of paper (not styro-save the enviro) bowls and some fork/spoons in that drawer, too. |
| | 4 | Title: quick and healthy; Content: My granddaughter loves these noodles (they're so yummy). They're ready in 4 minutes, and they're real noodles, unlike the ones that are just a combination of chemicals. | Title: Yummy and NO MSG!!; Content: This is one of the best instant ramen-bowls.<br />Great late night snack (N.B. cures hangovers), satisfies your hot & spicy cravings, is super fun to eat on camping and ski trips, and it has no MSG. (SO AWESOME.)<br />That said, beware the sodium content. Consuming the entire soup is an automatic 90% DV sodium intake. And honestly, you really don't need to use the whole packet to get the spice-kick and flavoring. Start with just half (or less, even!) and find your level of satisfaction. |
| | 5 | Title: pretty good; Content: These are quite tasty--soy sauce flavored but not too salty, which is misleading since it does contain plenty of salt( you probably shouldn't finish all of the soup, for health reasons). Another problem with these instant ramen is the lack of an expiration date on the package---I ordered Myojo's yakisoba along w/ this and had to throw most of it away. | Title: Not your everyday ramen (but it should be!); Content: I first picked up <a href="http://www.amazon.com/gp/product/B000LQOR DE">Nong Shim Shin Noodle Ramyun</a> in a local Asian Specialty store near my office. I needed a cheap lunch and this was my choice. Since that day I continued to buy it a couple packages at a time to keep at work for those days when I don't have a lunch planned. Now though I have an entire case because Amazon had such a great price. It's easy to cook up in the microwave and very filling. The one change I make is that I only use about 1/2 the seasoning pack, any more and it's too spicy for me to finish the entire bowl. Also I'm one of those "picky" people that doesn't like mushrooms and there are usually a couple small dried pieces in the vegetable pack that I pick out. Minor I know, BUT if you don't like mushrooms it's something you should know.<br /><br />While I still buy the "other" ramen for home because my kids like it, Nong Shim has won me over with theirs and it's my treat to myself. |

*Table 2*. The top five search results returned from our text-ada-002 vector database of 10,000 reviews when using all-mpnet-base-v2 + Vec2Vec versus text-ada-002 for the query "ramen noodles". Both models performed well and returned relevant results.

| Search Rank | Vec2Vec+ all-mpnet-base-v2 embedding results | text-ada-002 embedding results |
|---|---|---|
| 1 | Title: Wellness dog food; Content: My dog, a picky eater, loves this food. And it's made with better ingredients than most other stuff on the market. | Title: Good food - yum! Woof.; Content: My dog likes the Canidae, and he's very fussy. The fish product is a little, well, fishy, but the other varieties are solid. |
| 2 | Title: Dog won';t touch.; Content: Great ingredients! Really healthy. Good buy! Only problem. Even after I read him the ingredients, my dog won't touch them, even when he's hungry. :( | Title: Wellness dog food; Content: My dog, a picky eater, loves this food. And it's made with better ingredients than most other stuff on the market. |
| 3 | Title: Good; Content: This is good dog food, but I stopped buying it, because one of my dogs (I have two) did not like it. She is not a big fan of can food in general. | Title: Wellnes Dry Dog Food; Content: Good quality and our dogs will eat it. Many dry dog foods they just turn their noses up to. |
| 4 | Title: Good; Content: This is good dog food, but I stopped buying it, because one of my dogs (I have two) did not like it. She is not a big fan of can food in general. | Title: Dog food; Content: We have two 85-pound retriever mix dogs who really seem to enjoy the food. They give it five stars! The subscribe and save is a great deal for big bags of dog food--free delivery! |
| 5 | Title: Wellnes Dry Dog Food; Content: Good quality and our dogs will eat it. Many dry dog foods they just turn their noses up to. | Title: love this dog food; Content: We have been quite pleased with this dog food. Our dog has stopped itching and scratching, his coat is lustrous, and he eats less of it that other dog foods and seems to be maintaining his weight and being vigorous. |

*Table 3.* The top five search results returned from our text-ada-002 vector database of 10,000 reviews when using all-mpnet-base-v2 + Vec2Vec versus text-ada-002 for the query "dog food". Both models performed well and returned relevant results.

| Search Rank | Vec2Vec+ all-mpnet-base-v2 embedding results | text-ada-002 embedding results |
|---|---|---|
| 1 | Title: Good; Content: This is good dog food, but I stopped buying it, because one of my dogs (I have two) did not like it. She is not a big fan of can food in general. | Title: Expired; Content: The product was expired for a year before I received it. It was unusable, and I had to throw it away. |
| 2 | Title: Good; Content: This is good dog food, but I stopped buying it, because one of my dogs (I have two) did not like it. She is not a big fan of can food in general. | Title: Bad Cookies; Content: This product came in expired and of course stale. I purchased this for a school party and ended it using it as bird food. I was very disappointed. The only reason I didn't send it back was because I opened them before tasting it. I will never buy anything from this seller again. |
| 3 | Title: Dog won';t touch.; Content: Great ingredients! Really healthy. Good buy! Only problem. Even after I read him the ingredients, my dog won't touch them, even when he's hungry. :( | Title: expired !!!; Content: all the chocolate and everything was expired !!<br />this was a gift for my brother, and he did not eat any of it! |
| 4 | Title: Gave my dog the runs; Content: We have a 5 month old pit bull mix and she had very loose stool every time we tried to feed her this stuff. I offered the food to other people who have dogs and no one wanted it. I threw it away. What a waste. | Title: Watch expiration dates!1; Content: 1st: my cats really don't eat this brand with much relish like they did when Wellness had the mylar bags. We really miss those bags, the food just seemed to taste better to my cats. Canned, they won't touch any flavor but turkey.<br /><br />2nd: I've had 3 out of 12 in a case smell off and even the dogs won't touch it. So just check yours JIK.<br /><br />What has happened to this brand? They were the BEST 2 years ago! |
| 5 | Title: Was great, but formula changed!; Content: My dog was on Canidae for almost 2 years and she loved it. She was healthy with a full, shiny coat and plenty of energy. Then suddenly around December she started vomiting and having loose yellow stools, and became very lethargic. We took her to 2 vets and after several hundred dollars in tests they couldn't find anything wrong with her.<br /><br />I did some research and found that there were numerous people having the exact same problems I was! It turns out, they recently changed their formula, changed manufacturers, and decreased the amount of food in their package. Do some research, it's not the same high-quality food it used to be! | Title: Dented cans, stinky food; Content: Most of the cans were dented even though they were in the original case - makes me wonder if they were repackaged. The food is pretty stinky - the dog wouldn't eat it, and she loves cat food. However the cats scarfed it right down. |

*Table 4*. The top five search results returned from our text-ada-002 vector database of 10,000 reviews when using all-mpnet-base-v2 + Vec2Vec versus text-ada-002 for the query "the dog food was expired". Vec2Vec focused on "dog food" while text-ada-002 focused more on "expired".

| Search Rank | Vec2Vec+ all-mpnet-base-v2 embedding results | text-ada-002 embedding results |
|---|---|---|
| 1 | Title: very good oil; Content: item arrived on time, was of very high quality, and doesn't taste bitter like some extra virgins do. I'm very happy with my purchase. | Title: Fantastic delicious natural product delivered quickly; Content: I have actually visited the Pelindaba lavender fields a few years ago on a visit to the San Juan Islands so was thrilled to see this product available. I was reminded of the warm smell of lavender growing in the July sun with bees swarming around the plants. One of my best travel memories. Love this honey- it has a more lavender flavor than some honies but is really wonderful. Great value for the money. Adds a wonderful something extra to Greek yogurt or to a bowl of fresh blueberries. Fresh blueberries take very well to a little dollop of lavender honey. Try it! |
| 2 | Title: The "cashmere" of olive oils!; Content: My husband and I were served Olio Taibi "Biancolilla" at a dinner party tonight with rustic bread and fresh tomatoes - in a word...Amazing. This olive oil is flavorful, yet smooth and delicate. Contrary to the review above, I think Biancolilla would be fantastic in salad dressings and great for bread dipping. I look forward to giving bottles as gifts this holiday season! | Title: fine cat food; Content: It was delicious, better with a red wine than a white. The Garlic was a bit dominant over the sage.<br /><br />OK, it's cat food. I don't know what to say much except one thing:<br /><br />The strays I feed really like it.<br /><br />I'm not surprised, they like any food I give them, but canned more.<br /><br />It's a good value. So I'm pleased with it as a 'treat' for them. |
| 3 | Title: GREAT tea; Content: This tea, with its orange undertones, is marvelous. The aroma is great. Even non-tea drinkers will love this one. | Title: Pleasant soft lavender scent; Content: I Have burned about 60 sticks so far.<br /><br />Hem Lavender can be described as smooth, deep, and peaceful. It has a lavender scent but does not have the sharp edge that lavender usually carries. I do like the edge, therefore I give this item 4 not 5 stars. If you had not seen the package, you might not guess it was lavender, but it is a very nice scent without being overly floral.<br /><br />This can be described as an "all audiences" type of lavender. The males and females of my home enjoy this scent.<br />This scent is good enough to be part of my standard stock. I enjoy it in the morning since it is peaceful without being sleepy.<br /><br />If you are looking to add calm to your chaotic home, try this lavender. |
| 4 | Title: Very tasty!; Content: All organic ingredients in this very tasty blend. I just discovered this line of products last week, when a friend brought over two of the vinaigrettes, this one I'm reviewing and the Meyer lemon vinaigrette, which has an even brighter flavor.<br /><br />I drizzled this on shrimp and it was just lovely. It also works well on chicken and pork and, of course, salads. I really like the flavor it adds to a warm potato, sweet potato, carrot and eggs salad.<br /><br />It has a subtle garlic flavor, a nice touch of Italian seasonings and just the right amount of sea salt and black pepper.<br /><br />Very nice. | Title: Yum!; Content: this relaxing mate was delicious. It had rich bold flavors and that right touch of chocolate that warms a cold night. |
| 5 | Title: Superb taste; Content: I've tried a lot of oil oils over the years and this oil is full of flavor. I use it for cooking as well as tossing my fresh salads and vegetables. The value for the product is priced just right. The beautiful container makes for a nice look even if left out on the counter. I will purchase this product again. | Title: PERFECT AS I'd Hoped.; Content: I ordered this lavender about a month before it was scheduled to be delivered. Order was filled on time, with good communication. I was excited to open the box because the dried bunches are part of the decorations for our wedding and dinner/dance party afterwards. SO fragrant, well-preserved, evenly divided bunches! The bunches are of good size, with no scrimping on quality or amount. OF COURSE the whole box smells wonderful, too. Packing for shipping helped them arrive with 99.5% :) of buds intact. Highly recommend. |

*Table 5.* The top five search results returned from our text-ada-002 vector database of 10,000 reviews when using all-mpnet-base-v2 + Vec2Vec versus text-ada-002 for the query "the wine had notes of lavender". Vec2Vec returned several results related to oil instead of wine. Meanwhile, text-ada-002 seemed to focus on "lavender" but not "wine".

# Conclusion

In this paper, we introduced Vec2Vec, a lightweight neural network model capable of translating open-source MPNet embeddings into text-ada-002 embeddings. The model was trained and tested using a subset of 50,000 Amazon food reviews, providing an ample and diverse dataset of natural language. Our primary performance metric was cosine similarity between predicted and actual text-ada-002 embeddings. The Vec2Vec model achieved an impressive average cosine similarity of 0.932, indicating that the translation of embeddings was significantly accurate, given the high dimensionality of the target embedding space. However, we must emphasize that while the performance was commendable, the synthetic embeddings generated by our model could not fully match the performance of the actual text-ada-002 embeddings. This was observed in our manual quality assessments, where Vec2Vec fell short of text-ada-002's performance on complex vector search queries. Nonetheless, for simple queries, Vec2Vec's performance was close to the original model, demonstrating potential for its application in practical scenarios. Another limitation of our work is that Vec2Vec was only trained on food reviews. The model would most likely not perform as well on out-of-distribution data, such as news articles, tweets, or scientific writing. Also, different loss functions besides cosine similarity could be explored.

Given that the Vec2Vec model is lightweight, with a size less than 80 MB, and its ability to work offline, it could be a valuable asset in various contexts where either the API costs or the need for a constant internet connection can be prohibitive. As such, Vec2Vec could help democratize the use of vector databases by providing an offline and cost-effective alternative. Also, for sensitive data such as health information that is HIPAA-protected, using an offline model circumvents the need to send data to an external provider. Our research suggests that a larger, higher-performing embedding model could potentially be used to "teach" or "tune" a smaller model by adding a neural network on top.

Moving forward, there are several avenues for future work. Training a more sophisticated neural network, performing hyperparameter tuning, and leveraging larger datasets of paired embeddings could enhance performance and generalization capabilities. These datasets should include text from diverse domains, instead of just food reviews as was the case in this study. Additionally, incorporating a wider range of embedding models besides MPNet, such as BERT, Instructor, or RoBERTa, could extend the applicability of Vec2Vec. Finally, we believe that this work helps to pave the way for the development of tools that can robustly convert between different embedding spaces. We hope that Vec2Vec serves as a stepping stone for improved solutions aimed at enhancing interoperability, minimizing reliance on proprietary models, safeguarding data privacy, reducing costs, and allowing for offline operations.

The code, model weights, and dataset for testing are available on Hugging Face and Github.
Hugging Face: https://huggingface.co/gaodrew/vec2vec
Github: https://github.com/andrewgcodes/vec2vec

# References

[1] O. Levy and Y. Goldberg, "Dependency-Based Word Embeddings," Association for Computational Linguistics, 2014. Available: https://aclanthology.org/P14-2050.pdf

[2] M. Kusner, Y. Sun, N. Kolkin, and K. Weinberger, "From Word Embeddings To Document Distances," *proceedings.mlr.press*, Jun. 01, 2015. https://proceedings.mlr.press/v37/kusnerb15

[3] T. Mikolov, K. Chen, G. Corrado, and J. Dean, "Efficient Estimation of Word Representations in Vector Space," *arXiv.org*, Sep. 07, 2013. https://arxiv.org/abs/1301.3781

[4] J. Pennington, R. Socher, and C. Manning, "GloVe: Global Vectors for Word Representation," 2014. Available: https://nlp.stanford.edu/pubs/glove.pdf

[5] Y. Li and T. Yang, "Word Embedding for Understanding Natural Language: A Survey," *Studies in Big Data*, pp. 83–104, May 2017, doi: https://doi.org/10.1007/978-3-319-53817-4_4.

[6] N. Reimers, B. Schiller, T. Beck, J. Daxenberger, C. Stab, and I. Gurevych, "Classification and Clustering of Arguments with Contextualized Word Embeddings," *arXiv:1906.09821 [cs]*, Jun. 2019, Available: https://arxiv.org/abs/1906.09821

[7] J. Yao, Z. Dou, and J.-R. Wen, "Employing Personal Word Embeddings for Personalized Search," *Proceedings of the 43rd International ACM SIGIR Conference on Research and Development in Information Retrieval*, Jul. 2020, doi: https://doi.org/10.1145/3397271.3401153.

[8] K. Patel, D. Patel, M. Golakiya, P. Bhattacharyya, and N. Birari, "Adapting Pre-trained Word Embeddings For Use In Medical Coding," *ACLWeb*, Aug. 01, 2017. https://aclanthology.org/W17-2338/

[9] R. Greene, T. Sanders, L. Weng, and A. Neelakantan, "New and improved embedding model," *OpenAI*, Dec. 15, 2022. https://openai.com/blog/new-and-improved-embedding-model

[10] Sentence Transformers, "sentence-transformers/all-mpnet-base-v2 · Hugging Face," *huggingface.co*. https://huggingface.co/sentence-transformers/all-mpnet-base-v2

[11] K. Song, X. Tan, T. Qin, J. Lu, and T.-Y. Liu, "MPNet: Masked and Permuted Pre-training for Language Understanding," *arXiv.org*, Nov. 02, 2020. https://arxiv.org/abs/2004.09297 (accessed Jun. 20, 2023).

[12] J. J. McAuley and J. Leskovec, "From amateurs to connoisseurs," *Proceedings of the 22nd international conference on World Wide Web - WWW '13*, 2013, doi: https://doi.org/10.1145/2488388.2488466.

[13] A. K. Gao, "lightspeedEmbeddings," *GitHub*, Jun. 14, 2023. https://github.com/andrewgcodes/lightspeedEmbeddings (accessed Jun. 20, 2023).

[14] M. Abadi *et al.*, "TensorFlow: Large-Scale Machine Learning on Heterogeneous Distributed Systems," *arXiv.org*, 2016. https://arxiv.org/abs/1603.04467

[15] F. Chollet and others, "Keras: The Python Deep Learning library," *Astrophysics Source Code Library*, p. ascl:1806.022, Jun. 2018, Available: https://ui.adsabs.harvard.edu/abs/2018ascl.soft06022C/abstract